\pdfoutput=1

\documentclass[11pt]{article}

\usepackage[final]{acl}

\usepackage{times}
\usepackage{latexsym}

\usepackage[T1]{fontenc}

\usepackage[utf8]{inputenc}

\usepackage{microtype}

\usepackage{inconsolata}
\usepackage[para]{threeparttable}
\usepackage{booktabs}
\usepackage{multirow}
\usepackage{graphicx}
\usepackage{amsmath}
\usepackage{amssymb}
\usepackage{xcolor}
\usepackage{hyperref}
\usepackage{mathtools}
\usepackage[ruled,vlined]{algorithm2e}
\usepackage{tcolorbox}
\usepackage{listings}
\usepackage{enumitem}
\usepackage{makecell}
\definecolor{c1}{HTML}{0049C0}
\usepackage{colortbl}
\usepackage{tabularx}
\definecolor{mygray}{gray}{.95}
\definecolor{mycell}{rgb}{0.85, 0.93, 0.97}
\definecolor{mycelltwo}{RGB}{255, 238, 241}
\usepackage{longtable}
\usepackage{arydshln}
\tcbuselibrary{breakable}
\usepackage{amssymb}
\makeatletter
\renewcommand{\maketag@@@}[1]{\hbox{\m@th\normalsize\normalfont#1}}%
\makeatother

\newcommand{\ie}{\textit{i.e.}}
\newcommand{\eg}{\textit{e.g.}}

\usepackage{graphicx}
\usepackage{bbding}
\usepackage{algpseudocode}
\usepackage{tcolorbox}
\usepackage{needspace}
\tcbuselibrary{breakable} 
\tcbuselibrary{skins}

\newtcbox{\mybox}[1][]{
  enhanced,
  on line,     
  height=0.45cm,
  boxrule=0.8pt, 
  arc=0pt,     
  boxsep=1pt,  
  left=1pt, right=1pt, top=1pt, bottom=1pt, 
  colback=white, 
  coltext=gray,
  colframe=gray,
  valign=center,
}

\newtcbox{\algobox}[1][]{
  enhanced,
  on line,     
  height=0.45cm,
  boxrule=0.8pt, 
  arc=0pt,     
  boxsep=1pt,  
  left=1pt, right=1pt, top=1pt, bottom=1pt, 
  colback=white, 
  coltext=blue!45,
  colframe=blue!45,
  valign=center,
}

\title{CRMWeaver: Building Powerful Business Agent via Agentic RL and Shared Memories}

\author{
Yilong Lai\textsuperscript{1,2}\thanks{\ Work done during interned at Alibaba Group.},\hspace{1.0mm}
    Yipin Yang\textsuperscript{1},\hspace{1.0mm}
   Ting Liang\textsuperscript{1},\hspace{1.0mm}
   Jialong Wu\textsuperscript{2},\hspace{1.0mm}
   Zhenglin Wang\textsuperscript{2},\hspace{1.0mm}\\
   \textbf{Jianguo Lin}\textsuperscript{1}\textbf{,}\hspace{1.0mm}
   \textbf{Keping Yang}\textsuperscript{1}\thanks{~~Corresponding Author.}\hspace{1.5mm}\\
    \textsuperscript{1} Taobao\ \&\ Tmall Group of Alibaba \quad
    \textsuperscript{2} Southeast University \\
\texttt{laiyilong0@gmail.com, \{uipin.yang, kuiyu.lt, tengxiao, shaoyao\}@alibaba-inc.com}
}

\begin{document}
\maketitle
\begin{abstract}
Recent years have seen the rapid development of LLM-based agents, which shed light on the use of language agents to solve complex real-world problems.
A prominent application lies in business agents, which interact with databases and internal knowledge bases via tool calls to fulfill diverse user requirements. 
However, this domain is characterized by intricate data relationships and a wide range of heterogeneous tasks, from statistical data queries to knowledge-based question-answering. 

To address these challenges, we propose CRMWeaver, a novel approach that enhances business agents in such complex settings. 
To acclimate the agentic model to intricate business environments, we employ a synthesis data generation and RL-based paradigm during training, which significantly improves the model's ability to handle complex data and varied tasks.
During inference, a shared memories mechanism is introduced, prompting the agent to learn from task guidelines shared by similar problems, thereby further boosting its effectiveness and generalization, especially in unseen scenarios.
We validate the efficacy of our approach on the CRMArena-Pro dataset, where our lightweight model achieves competitive results in both B2B and B2C business scenarios, underscoring its practical value for real-world applications.
\end{abstract}

\section{Introduction}
With the advanced reasoning and decision-making capabilities, agentic systems based on large language models (LLMs)~\cite{weng2023agent, yao2023react} are poised to advance artificial intelligence toward more human-like intelligence by autonomously interacting with external environments to accomplish complex goals.
A large number of tasks have proven to be effective using an agent-based framework, including coding~\cite{jimenez2023swe, wang2025openhands}, web browsing~\cite {zhouwebarena, deng2023mind2web}, science discovery~\cite{zhao2025sciarena, xiang2025scireplicate}, and deep research~\cite{wu2025webdancer, li2025websailor}.

\begin{figure}[t]
    \centering
    \includegraphics[width=1\linewidth]{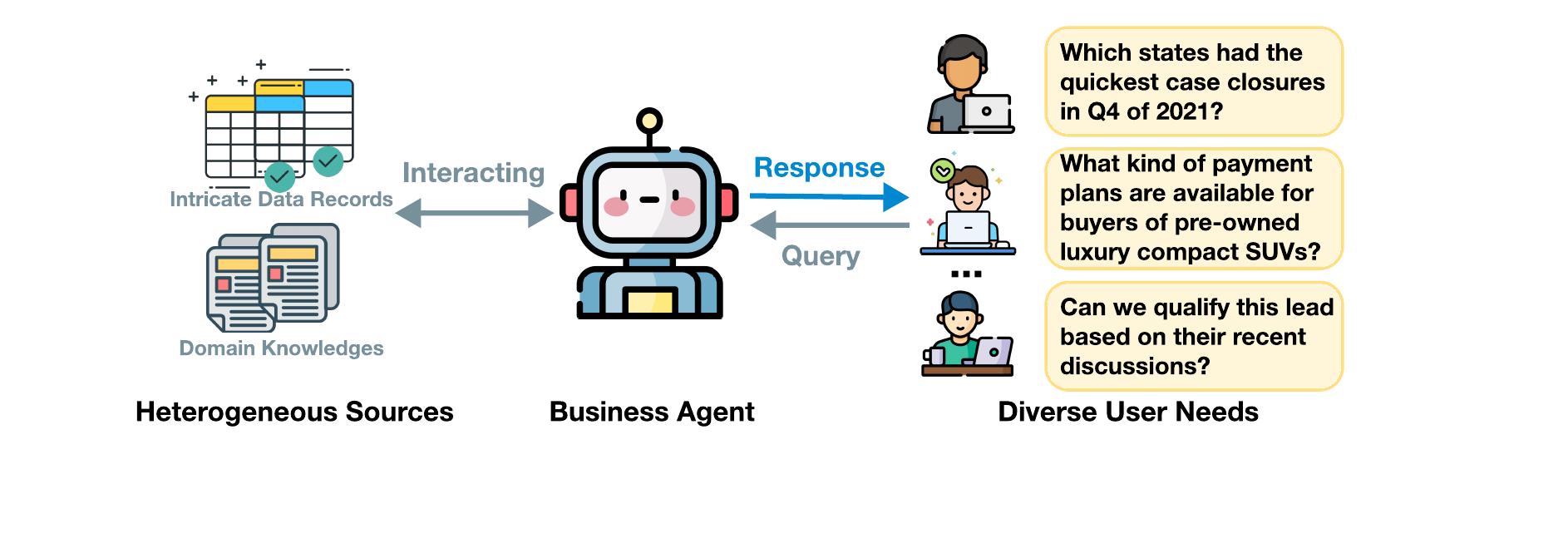}
    \caption{Challenges of business agents in handling diverse user needs and interacting with heterogeneous data sources.} 
    \label{fig:crmweaver_overview}
\end{figure}

Business agents constitute a prominent application scenario, where systems interact with enterprise databases and internal knowledge bases through tool calls to satisfy diverse user requirements. 
Unlike general-purpose agents, this domain is marked by highly intricate data dependencies (e.g., large-scale interconnected tables) and task heterogeneity that spans from statistical data queries to knowledge-based question answering.
Figure~\ref{fig:crmweaver_overview} depicts the challenges of the business agent.
As a result, only proficient but prohibitively costly models have been able to achieve relatively satisfactory performance in these contexts.

To address these challenges, we propose \textbf{CRMWeaver}, a novel framework for building robust business agents through a two-fold optimization strategy. 
First, to adapt the model to the intricacies of business environments while mitigating the scarcity of high-quality training data, we employ large language models to synthesize complex and diverse training data. 
Based on the synthesis data, we introduce a two-stage training paradigm. 
In the first stage, we utilize rejection sampling and trajectory distillation to bootstrap the model, enabling it to solve the queries through multi-turn interaction. 
In the second stage, we employ reinforcement learning with Decoupled Clip and Dynamic Sampling Policy Optimization(DAPO)~\cite{yu2025dapo} to further enhance the model’s generalization in complex business environments\footnote{To rigorously assess generalizability while mitigating data leakage, we conduct model training on CRMArena and evaluation on CRMArena-Pro, which encompass entirely distinct environments.}.
Second, to cope with the diversity of user queries and the continual evolution of business schemas, we introduce a long-term memory module to improve the agent’s stability across multiple tasks, particularly for previously unseen tasks. 
Specifically, we construct an index of task-specific guidelines built from successful trajectories. 
At inference time, when similar tasks are identified, relevant memories are retrieved and injected into the current context, enabling the agent to leverage prior knowledge and improve task performance.

Our contributions are threefold:
\begin{enumerate}
    \item We address the scarcity of high-quality training data and the need to adapt to complex business environments by combining synthetic data generation with a two-stage training pipeline.
    \item We introduce a long-term memory module that enhances robustness and generalization by efficiently reusing solution guidelines from a stronger reasoning model, enhancing the agent’s capabilities on diverse, complex tasks.
    \item A lightweight model built on Qwen3-4B achieves performance comparable to Qwen3-235B-Instruct and Gemini 2.5-Pro on the business-agent benchmark CRMArena-Pro across B2B and B2C scenarios, demonstrating the effectiveness of CRMWeaver.
\end{enumerate}

\section{Preliminaries}~\label{sec:preliminary}
CRMWeaver is built and evaluated on CRMArena and CRMArena-Pro~\cite{huang-etal-2025-crmarena,huang2025crmarena-pro}, two realistic benchmarks for business-oriented LLM agents.
CRMArena simulates a Salesforce-based customer service environment with 16 interconnected objects and latent causal dependencies, supporting nine expert-designed service tasks.
CRMArena-Pro further scales this setting to 25 Salesforce objects across Service, Sales, and Configure-Price-Quote schemas, covering over 80K records and nineteen tasks that span database querying, textual reasoning, workflow execution, and policy compliance.
Together, these benchmarks require agents to reason over highly interconnected enterprise data and solve complex tasks through multi-turn interactions.

We formulate task solving as a standard agent–environment interaction process.
At each time step $t$, the agent observes $o_t$ and selects an action $a_t$ based on the interaction history $c_t$.
This process is modeled as a Partially Observable Markov Decision Process (POMDP) $\langle \mathcal{U}, \mathcal{S}, \mathcal{A}, \mathcal{O}, \mathcal{T}, \mathcal{R} \rangle$, where $\mathcal{U}$ denotes user queries, $\mathcal{S}$ denotes latent environment states, and $\mathcal{R}$ indicates task completion.
The action space $\mathcal{A}$ consists of three tools: (i) \textbf{\textit{Execute}} for querying records via SQL or Salesforce Object Search Language(SOSL), (ii) \textbf{\textit{Date Calculation}} for temporal reasoning, and (iii) \textbf{\textit{Answer}} for submitting the final response.

\begin{figure*}[t]
    \centering
    \includegraphics[width=1.0\textwidth]{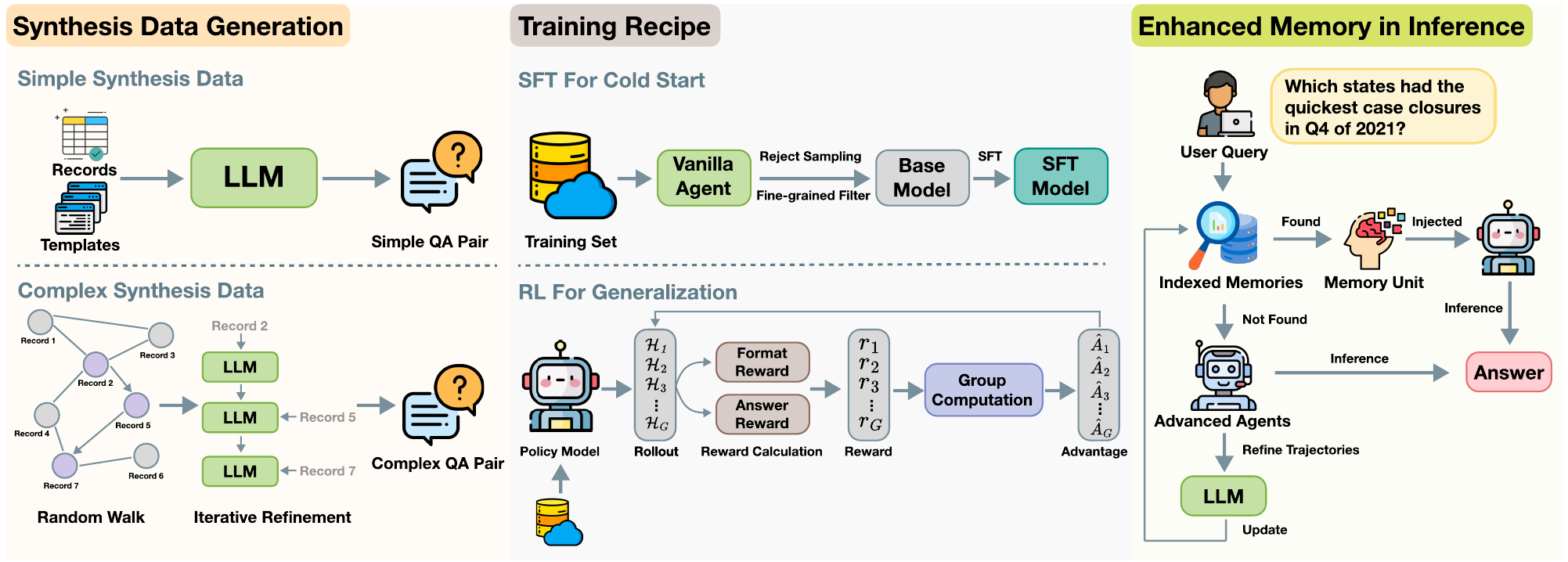}
    \caption{The overall pipeline of CRMWeaver. We first construct both simple and complex synthetic data by leveraging correlations among data records to support model training (\S~\ref{sec:data}). The training recipe adopts a two-stage paradigm, including SFT for model initialization and reinforcement learning for generalization (\S~\ref{sec:training}). Finally, the inference process incorporates a memory enhancement mechanism (\S~\ref{sec:memory}).} 
    \label{fig:overview}
\end{figure*}

The above setup defines the environments, tasks, and interaction paradigm that ground our work. 
We next introduce CRMWeaver, outlining its methodology for enhancing the problem-solving business-oriented tasks.

\section{Methodology}
This section introduces CRMWeaver, including data synthesis for cold-start training(\S~\ref{sec:data}), the agent framework(\S~\ref{sec:agent_setup}), and a two-stage training pipeline(\S~\ref{sec:training}) with SFT and DAPO-based reinforcement learning.
We further integrate long-term memory(\S~\ref{sec:memory}) at inference to improve robustness in complex and unseen tasks.
Figure~\ref{fig:overview} illustrates the overall framework.

\subsection{Synthesis Data Construction}\label{sec:data}
Given that CRMArena provides only a limited set of task-specific test queries targeting particular objects, we aim to enhance the model's generalization and reasoning capabilities over complex, multi-relational, and heterogeneous data. 
Inspired by recent work leveraging synthetic complex QA datasets to improve deep research agents~\cite{wu2025webdancer, li2025websailor, shi2025taskcraft}, we construct a graph $G$, where nodes correspond to records and edges represent inter-table relationships, thereby capturing the relational structure inherent in CRMArena.
For each synthesis data sample $X = (Q,A)$, we first extract a path $\Psi$ through a random walk on the graph $G$:
\begin{equation}\label{eq:random_walk}
    \Psi = (e_1, e_2, \cdots , e_k) \sim \operatorname{RandomWalk}(G)
\end{equation}
where $Q$ is the synthesis query, $A$ is the corresponding answer, $e_i$ is the sampled record in the path $\Psi$, and $k$ is the length of the sample path $\Psi$.

For each pair $\{e_{i-1}, e_i\}$ along the path $\Psi$, the connection is established through a specific field both in the records $e_{i-1}$ and $e_i$. 
Starting from the initial node $e_1$, we take the unique identifier of $e_1$ as the final answer $A$. 
We then define $q_i$ as a synthetic sub-query, derived from the partial path $(e_1, e_2, \cdots, e_i)$. 
To construct these sub-queries, we employ a prompt-based LLM strategy, where the LLM generates a more complex query $q_i$ given the preceding query $q_{i-1}$ and the corresponding record pair $\{e_{i-1}, e_i\}$:
\begin{equation}\label{eq:data_generation}
    q_i = \operatorname{LLM}(\{e_{i-1}, e_i\} | q_{i-1}, A)
\end{equation}

After $k$ iterations, we obtain a high-quality synthetic data $X = (Q, A)$. 
In addition, to broaden the scope and diversity of training tasks, we further construct synthetic datasets targeting statistical analysis and knowledge-base question answering, by exploiting tabular statistical features and internal knowledge resources.  
The detailed prompts used for constructing synthesis data are provided in the Appendix~\ref{appendix:synthesis_data}.

Overall, our training corpus encompasses three categories:
\begin{itemize}
\item \textit{\underline{Complex synthetic data}}: Crafted using the method in Eq.~\eqref{eq:random_walk} and Eq.~\eqref{eq:data_generation} to strengthen the model’s ability to search, integrate, and reason over multi-relational queries across interconnected tables. 
We provide some examples of complex synthesis queries in Appendix~\ref{appendix:complex_synthesis_data_example}.
\item \textit{\underline{Simple synthetic data}}: Designed to instill basic business QA and data analysis skills, serving as a foundation for business problem-solving. (\eg How long after purchase can Shoes \& Clothings customers claim store credit for an overcharged order?)
\item \textit{\underline{Task-specific data}}: Consisting of nine well-defined tasks from CRMArena. (\eg Which states had the quickest case closures in Q4 of 2021?)
\end{itemize}

\subsection{Agent Setup of CRMWeaver}\label{sec:agent_setup}
We build CRMWeaver on top of the \textbf{\textit{ReAct}} framework~\cite{yao2023react}, aligning with prior works~\cite{huang-etal-2025-crmarena, huang2025crmarena-pro}. 
Formally, given a query $Q$, the ReAct agent addresses the task through iterative cycles conditioned on the action space $\mathcal{A}$ and observation space $\mathcal{S}$. 
At time step $t$, with historical trajectory $\mathcal{H}_t = (Q,  \tau_1, \alpha_1, o_1, \cdots, \tau_{t-1}, \alpha_{t-1}, o_{t-1})$, the agent derives the current thought $\tau_t$ and action $\alpha_t$ according to the following equation:
\begin{equation}
    (\tau_t, \alpha_t) = \pi_\theta(\tau_t, \alpha_t | \mathcal{H}_t)
\end{equation}
where $\pi_\theta$ is the policy of the model $\theta$, and $\alpha_t \in \mathcal{A}$. 
As mentioned in Section~\ref{sec:preliminary}, the Action Space $\mathcal{A}$ comprises three tools: \textbf{\textit{execute}}, \textbf{\textit{date calculation}}, and \textbf{\textit{answer}}, which enable the agent to interact with external business environments, perform precise temporal computations, and submit the final answers.
We provide a detailed trajectories example of CRMWeaver in Appendix~\ref{appendix:case_study}.


\subsection{Two-stage Training Recipe}\label{sec:training}
\subsubsection{SFT for Agentic Model Cold Starts}

To equip the model with the fundamental capability to address business tasks and ensure it follows the \textbf{\textit{ReAct}} paradigm during problem-solving, we first leverage a powerful language model, GPT-4.1.
Reject Sampling~\cite{yuan2023scalingrelationshiplearningmathematical, grattafiori2024llama3herdmodels} is applied to the training data mentioned in Section~\ref{sec:data}, generating multiple candidate execution trajectories.
Subsequently, we filter these trajectories based on two primary criteria:
\begin{itemize}[itemsep=0pt, topsep=0pt]
    \item \textbf{Correctness of results}: To ensure the accuracy of the trajectory data, we first exclude trajectories whose final answer does not match the gold answer.
    \item \textbf{Correctness of process}: A portion of the training data has ``None'' as the answer. To ensure the model does not arrive at the result through simple guesswork, we discard trajectories with a very short number of execution turns.
\end{itemize}

This process yields a high-quality training dataset, $\mathcal{D}_{\text{SFT}}$, which contains a large number of execution trajectories. 
Each sample in the dataset is a high-quality trajectory $\mathcal{H} = (Q, \tau_1, \alpha_1, o_1, \cdots, \tau_t, \alpha_t, o_t)$.
Following this, we train the model $\theta$ using supervised fine-tuning (SFT). 

\subsubsection{Reinforcement Learning to Generalization}
To further enhance the agentic model's capabilities, particularly its generalization in complex business environments, we employed a training strategy based on Reinforcement Learning, which allows the model to refine its policy through multi-turn interactions with an external environment.

In details, for each sample $(Q,A)$ from our training dataset $\mathcal{D}$, we rollout a group of trajectories, denoted as $\{\mathcal{H}_i\}_{i=1}^G$.
Subsequently, we optimize the policy of model $\theta$ through Decouple Clip and Dynamic Sampling Policy (DAPO)~\cite{yu2025dapo}:
\begin{equation}\small
\hspace{-0.8cm}
\begin{aligned}
\mathcal{J}_{\text{RL}}(\theta) 
   &= \mathbb{E}_{\substack{(Q, A) \sim \mathcal{D} \\ 
        \{\mathcal{H}_i\}_{i=1}^G \sim \pi_{\theta_{\text{old}}}(\cdot \mid q)}} \Bigg[
       \frac{1}{\sum_{i=1}^G |\mathcal{H}_i|} 
       \sum_{i=1}^G \sum_{t=1}^{|\mathcal{H}_i|} \\
   &\min\!\Big(
         r_{i,t}(\theta)\,\hat{A}_{i,t},\;
         \operatorname{clip}\!\big(r_{i,t}(\theta),\, 
            1-\varepsilon_{\text{low}},\, 
            1+\varepsilon_{\text{high}}\big)\hat{A}_{i,t}
       \Big)
     \Bigg] \\
   &\text{s.t.}\quad 
      0 < \big|\{\mathcal{H}_i \mid \mathbf{is\_equivalent}(A, \mathcal{H}_i)\}\big| < G
\end{aligned}
\end{equation}
where
\begin{equation}
\begin{aligned}
r_{i,t}(\theta) 
   &= \frac{\pi_\theta(\mathcal{H}_{i,t}\mid Q, \mathcal{H}_{i,<t})}
           {\pi_{\theta_{\text{old}}}(\mathcal{H}_{i,t}\mid Q, \mathcal{H}_{i,<t})}, \\[6pt]
\hat{A}_{i,t} 
   &= \frac{R_i - \operatorname{mean}\!\big(\{R_i\}_{i=1}^G\big)}
           {\operatorname{std}\!\big(\{R_i\}_{i=1}^G\big)} .
\end{aligned}
\end{equation}

We use a rule-based reward function $R$, defined as follows:
\begin{equation}
    R(\hat{a}_i, a) = 0.1 \times s_{\text{format}} + 0.9 \times s_{\text{answer}} 
\end{equation}

where the $s_{\text{format}}$ is the format score, which is awarded to the model $\theta$ only when its output contains complete and valid pairs of \mybox{<think>}...\mybox{</think>}, \mybox{<tool\_call>}...\mybox{</tool\_call>} tags, and ended with a valid \mybox{<answer>}...\mybox{</answer>} tag.
The $s_{\text{answer}}$ is the answer score. 
For data analysis tasks, this reward is given only if the final answer $\hat{a}_i$ exactly matches the gold answer $a$. For QA tasks, we use the F1 score as the answer reward.

\begin{table*}[t]
    \centering
    \small
    \begin{threeparttable}
    \resizebox{2.0\columnwidth}{!}{%
    \begin{tabular}{lccccc|ccccc}
        \toprule
        & \multicolumn{5}{c|}{\textbf{B2B}} & \multicolumn{5}{c}{\textbf{B2C}} \\
        \textbf{Model} & \textbf{Workflow} & \textbf{Policy} & \textbf{Text} & \textbf{Database} & \textbf{Avg} & \textbf{Workflow} & \textbf{Policy} & \textbf{Text} & \textbf{Database} & \textbf{Avg} \\
        \midrule
        GPT-4o-mini & 19.5 & 33.8 & 12.6 & 19.3 & 21.3 & 13.5 & 25.6 & 11.7 & 23.8 & 18.6 \\
        GPT-4o & 26.0 & 27.5 & 22.1 & 31.3 & 26.7 & 29.0 & 32.5 & 22.2 & 33.1 & 29.2 \\
        Gemini-2.5-Flash & 67.5 & 33.5 & 25.1 & 41.6 & 41.9 & 80.0 & 31.0 & 26.3 & 49.3 & 46.7 \\
        OpenAI-o1 & 67.0 & 43.3 & 23.4 & 56.3 & 47.5 & 74.5 & 34.5 & 29.5 & 59.6 & 49.5\\
        Gemini-2.5-Pro & 83.0 & 41.0 & 34.7 & 57.6 & 54.1 & 90.0 & 42.0 & 36.2 & 64.9 & \textbf{58.3} \\
        Qwen3-235B-Instruct & 90.0 & 37.7 & 30.1 & 49.8 & 51.9 & 87.5 & 42.0 & 33.5 & 50.0 & 53.3 \\
        Kimi-K2-Instruct & 92.0 & 42.5 & 34.5 & 48.0 & \underline{54.3} & 93.0 & 41.8 & 36.2 & 52.2 & 55.8 \\ 
        \midrule
        \textbf{CRMWeaver\textit{(Ours)}} & 90.0 & 27.3 & 32.1 & 73.0 & \textbf{55.6} & 90.5 & 31.3 & 33.8 & 72.8 & \underline{57.1} \\
        \bottomrule
    \end{tabular}
    }
    \end{threeparttable}
    \caption{
    Experiments result on CRMArena-Pro across the B2B and B2C domains.
    The best average results among all backbones are \textbf{bolded}, and the second-best results are \underline{underlined}.
    }
    \label{table:main_exp}
\end{table*}

\subsection{Improving Reasoning with Long-term Memory}\label{sec:memory}

For business-oriented tasks in CRMArena and CRMArena-Pro, we find that diverse task surface forms often correspond to shared latent solution structures.
For example, queries like \textit{``Find the agent with the shortest handle time who managed more than one case in Winter 2021''} and \textit{``In the past three months, which agent achieved the lowest average handle time while handling more than two cases?''} differ linguistically but follow highly similar workflow patterns.
At the same time, business tasks are inherently diverse and frequently involve previously unseen compositions of constraints, posing significant challenges for generalization.

To address this, we introduce a long-term memory module that augments the agent's inference process with reusable workflow guidelines, inspired by prior work on long-term memory for agentic systems~\cite{tang2025agent, fang2025memp}.
As illustrated in Algorithm~\ref{algorithm:memory} in Appendix~\ref{appendix:long-term-memory}, the module comprises an \textbf{\textit{online memory-guided inference}} phase and an \textbf{\textit{offline memory consolidation}} phase.

In the online phase, the agent retrieves the most relevant memory unit from an indexed memory pool $\mathcal{M}$ given a query $Q$.
When a relevant memory is retrieved, it is incorporated into the model's contextual input, allowing the agent to reuse empirically validated solution patterns for analogous tasks.

For queries without sufficiently similar precedents, the offline consolidation phase is invoked.
A more powerful consolidation model resolves the query to produce execution trajectories, from which a high-level workflow guideline is extracted as a candidate memory unit.
A verification step is then applied to determine whether the generated guideline should be added to the indexed memory pool, ensuring the quality of stored memories.

Formally, a memory unit $\mathcal{E}$ is constructed by converting the consolidation trajectory $\tilde{\mathcal{S}}$ into an abstract guideline using an LLM:
\begin{equation}
\mathcal{E} = \operatorname{LLM}(\tilde{\mathcal{S}}, Q).
\end{equation}
Examples and implementation details are provided in Appendix~\ref{appendix:memory_generation}.

\section{Experiments}
\subsection{Experiments Setup}
\textbf{Implementation Details} \quad The backbone for our model $\theta$ is \texttt{Qwen3-4B-Instruct-2507}. 
The retrieval model used in \S~\ref{sec:memory} is \texttt{BGE-small-en-v1.5} and the advanced model is \texttt{o3-mini}.
We used the llama-factory framework~\cite{zheng2024llamafactory} for Supervised Fine-Tuning (SFT) on nearly 3k trajectories.
In Stage 2, we perform reinforcement learning using the verl\footnote{\url{https://github.com/volcengine/verl}} framework on 3.5k question–answer pairs.
And all the results are obtained under the standard \textbf{\textit{ReAct}} paradigm.
During evaluation, the memory module is updated within a single run following the predefined consolidation rules, as described in the Appendix~\ref{appendix:evaluate_protocol}.
We provide more implementation details in Appendix~\ref{appendix:implementation_details}.

\noindent\textbf{Datasets Details} \quad As mentioned in \S\ref{sec:preliminary}, we use CRMArena-Pro~\cite{huang2025crmarena-pro} as the testbed of our proposed method, which is composed of nineteen distinct, expert-validated tasks categorized into four core business skills: Database Querying \& Numerical Computation\textbf{(Data)}, Information Retrieval \& Textual Reasoning\textbf{(Text)}, Workflow Execution\textbf{(Workflow)}, and Policy Compliance\textbf{(Policy)}.
The dataset encompasses both Business-to-Business(B2B) and Business-to-Consumer(B2C) scenarios. 
We adopted the single-turn setting, \ie the user provides the task at once, resulting in a total of 3.8k test data samples.
Evaluation in CRMArena-Pro is multifaceted. 
For sub-tasks such as knowledge QA, we employed the F1 score as the evaluation metric, following standard settings.
For other tasks, we used exact match as the metric.
Details and statistics of CRMArena-Pro are provided in Appendix~\ref{appendix:crmarena-details}.

\subsection{Main Experiments}


As shown in Table \ref{table:main_exp}, we present a comprehensive evaluation of our proposed method CRMWeaver against a suite of robust baseline models. 
The performance assessment was conducted on the CRMArena-Pro dataset, which covers both B2B and B2C domains and consists of 19 distinct subtasks.

Our model demonstrates superior performance, achieving a highly competitive average score of 55.6 in the B2B domain and a score of 57.1 in the B2C domain compared with much more advanced LLMs. 
Notably, CRMWeaver exhibits a pronounced advantage in database-related tasks, where it outperforms all baselines with scores of 73.0 (B2B) and 72.8 (B2C).
The results across diverse domains and environments demonstrate the effectiveness and generalization capability of the proposed method.

\section{Analysis}
\subsection{Ablation Studies for CRMWeaver}
\begin{table}[t]
    \centering
    \small
    \begin{threeparttable}
    \begin{tabular*}{0.98\columnwidth}{lccc}
        \toprule
        \textbf{Configuration} & \textbf{B2B(Avg)} & \textbf{B2C(Avg)}\\
        \midrule 
        \textcolor{gray}{\textit{Backbone Only}} \\
        Qwen3-4B-Instruct & 19.9 & 20.3 \\
        \midrule
        \textcolor{gray}{\textit{CRMWeaver Variants}} \\
        \textbf{Full\ (SFT + RL + Memory)} & \textbf{55.6} & \textbf{57.1} \\
        \quad w/o\hspace{1mm} Long-term Memory & 54.5 & 55.3 \\
        \quad w/o\hspace{1mm} RL Training & 49.2 & 48.5 \\
        \quad w/o\hspace{1mm} SFT + RL & 23.7 & 24.5 \\
        \bottomrule
    \end{tabular*}
    \end{threeparttable}
    \caption{
    Ablation studies for CRMWeaver on the CRMArena-Pro benchmark.
    }
    \vspace{-4mm}
    \label{table:ablation}
\end{table}

To examine the contribution of each component, we conduct ablation studies on our proposed framework, as shown in Table~\ref{table:ablation}.
Compared to directly using the backbone model, our method yields notable improvements on CRMArena-Pro, achieving average score gains of \textbf{179.4\%} (from 19.9 to 55.6) in the B2B domain and \textbf{181.2\%} (from 20.3 to 57.1) in the B2C domain.

In addition, we analyze the effect of individual design choices in CRMWeaver by removing each stage in turn. 
In all cases, excluding a component results in a decrease in performance, indicating that each stage contributes to the overall results.

\subsection{The Frequency of Updates to Long-Term Memory}
Our analysis indicates that updates to the Long-term Memory module are relatively infrequent, whereas its indexed memories exhibit consistently high utilization across a broad range of tasks.

In the CRMArena-Pro B2B scenario, the average update frequency is close to 3\%, indicating that only 3 out of 100 business tasks, on average, trigger an indexed memories update. 
We observe that, for a small subset of tasks, such as Knowledge QA, low question similarity leads to a low probability of accessing the indexed memory. 
This can lead to frequent updates. 
To mitigate this, we employ a few-shot prompting approach with an LLM to classify whether a given query constitutes a knowledge-based question. This mechanism effectively prevents memory updates for QA tasks, thereby avoiding an excessive update frequency.

\section{Related Works}
\textbf{Business Agents and Benchmarks.} \quad
Evaluation of business agents has evolved from synthetic settings to realistic enterprise environments.
Early benchmarks such as WebShop~\cite{yao2022webshop} relied on simplified webpages, while later efforts, including WorkArena~\cite{drouin2024workarena}, WorkBench~\cite{workbench}, and Tau-bench~\cite{yao2024tau}, introduced outcome-oriented tasks on enterprise platforms.
CRMArena~\cite{huang-etal-2025-crmarena} significantly advanced realism by modeling Salesforce-based CRM environments with expert-verified tasks across multiple interconnected objects.
CRMArena-Pro~\cite{huang2025crmarena-pro} further expanded this setting in both scale and task diversity, providing the most comprehensive evaluation for business-oriented agents to date.

\noindent\textbf{Long-term Memory of Language Agent.} \quad
Long-term memory has been widely studied to improve agent performance by leveraging historical trajectories and accumulated knowledge~\cite{long-term-memory-survey,mo2025survey}.
Existing approaches explore different memory representations, including human-inspired memory dynamics~\cite{zhong2024memorybank}, latent-space knowledge encoding~\cite{memoryllm}, and structured or reusable memories for cross-task generalization~\cite{agent-workflow-memory, tang2025agent}.
More recent work, such as Mem0~\cite{chhikara2025mem0}, improves scalability via graph-based dynamic memories that capture complex relational dependencies beyond fixed context windows.

\noindent\textbf{Agentic Reinforcement Learning.} \quad
Agentic reinforcement learning enhances LLMs through multi-turn interactions in complex environments.
Rule-based RL has gained prominence with models such as Deepseek-R1~\cite{guo2025deepseek}.
Prior work applies agentic RL to retrieval and tool-augmented reasoning, including Search-R1~\cite{jin2025search} and ReTool~\cite{feng2025retool}.
Recent systems such as WebDancer, WebSailor, and WebShaper~\cite{wu2025webdancer, li2025websailor, tao2025webshaper} further demonstrate the effectiveness of data synthesis and end-to-end agentic RL in large-scale, real-world web interaction tasks.

\section{Conclusion}
We present CRMWeaver, a framework for building effective business agents in complex enterprise environments.
CRMWeaver addresses data scarcity and domain complexity through synthesized training data, a two-stage learning pipeline that combines supervised initialization with DAPO-based reinforcement learning, and a long-term memory mechanism that enables knowledge reuse at inference.

Together, these components provide a practical and cost-efficient approach for deploying reliable business agents across diverse and evolving real-world tasks.

\section*{Limitation}
\textbf{Multi-turns Settings.} \quad Due to constraints in both training frameworks and implementation techniques, our reinforcement learning setup primarily focuses on scenarios where the user provides a single query and the agent interacts with tools through multiple tool call/observation turns. 
We have not yet extended our framework to more complex multi-turn user interactions~\cite{mo2023convgqr, adacqr, adarewriter} as explored in environments such as CRMArena-Pro~\cite{huang2025crmarena-pro} and Tau-bench~\cite{yao2024tau}. 

\noindent\textbf{GPU Utilization During Rollout.} Although we adopt asynchronous rollout optimization, the rollout process remains time-consuming and often leads to underutilized GPU resources. 
As a result, we are forced to restrict both the context length and the number of tool calls. 
We plan to address this limitation in future work by leveraging fully asynchronous RL frameworks such as AReal~\cite{areal-rl-framework}. 

\noindent\textbf{Context Management} \quad To handle the common issue of context window limitations during training and inference, we impose a simple constraint on the number of records returned, which inevitably restricts the agent’s reasoning capacity. 
We envision adopting more sophisticated methods such as ReSum~\cite{resum}, to manage contextual information more effectively. 

\noindent\textbf{Hardware} \quad Due to hardware constraints, we are unable to conduct agent RL training on larger-scale models (e.g., 14B or 32B parameters).

\bibliography{custom}

@article{weng2023agent,
  title   = "LLM-powered Autonomous Agents",
  author  = "Weng, Lilian",
  journal = "lilianweng.github.io",
  year    = "2023",
  month   = "Jun",
  url     = "https://lilianweng.github.io/posts/2023-06-23-agent/"
}

@inproceedings{yao2023react,
  title={React: Synergizing reasoning and acting in language models},
  author={Yao, Shunyu and Zhao, Jeffrey and Yu, Dian and Du, Nan and Shafran, Izhak and Narasimhan, Karthik and Cao, Yuan},
  booktitle={International Conference on Learning Representations (ICLR)},
  year={2023}
}

@article{jimenez2023swe,
  title={Swe-bench: Can language models resolve real-world github issues?},
  author={Jimenez, Carlos E and Yang, John and Wettig, Alexander and Yao, Shunyu and Pei, Kexin and Press, Ofir and Narasimhan, Karthik},
  journal={arXiv preprint arXiv:2310.06770},
  year={2023}
}

@inproceedings{zhouwebarena,
  title={WebArena: A Realistic Web Environment for Building Autonomous Agents},
  author={Zhou, Shuyan and Xu, Frank F and Zhu, Hao and Zhou, Xuhui and Lo, Robert and Sridhar, Abishek and Cheng, Xianyi and Ou, Tianyue and Bisk, Yonatan and Fried, Daniel and others},
  booktitle={The Twelfth International Conference on Learning Representations}
}

@article{wu2025webdancer,
  title={WebDancer: Towards Autonomous Information Seeking Agency},
  author={Wu, Jialong and Li, Baixuan and Fang, Runnan and Yin, Wenbiao and Zhang, Liwen and Tao, Zhengwei and Zhang, Dingchu and Xi, Zekun and Fu, Gang and Jiang, Yong and others},
  journal={arXiv preprint arXiv:2505.22648},
  year={2025}
}

@article{li2025websailor,
  title={WebSailor: Navigating Super-human Reasoning for Web Agent},
  author={Li, Kuan and Zhang, Zhongwang and Yin, Huifeng and Zhang, Liwen and Ou, Litu and Wu, Jialong and Yin, Wenbiao and Li, Baixuan and Tao, Zhengwei and Wang, Xinyu and others},
  journal={arXiv preprint arXiv:2507.02592},
  year={2025}
}

@article{xiang2025scireplicate,
  title={Scireplicate-bench: Benchmarking llms in agent-driven algorithmic reproduction from research papers},
  author={Xiang, Yanzheng and Yan, Hanqi and Ouyang, Shuyin and Gui, Lin and He, Yulan},
  journal={arXiv preprint arXiv:2504.00255},
  year={2025}
}

@article{mo2025survey,
  title={A survey of conversational search},
  author={Mo, Fengran and Mao, Kelong and Zhao, Ziliang and Qian, Hongjin and Chen, Haonan and Cheng, Yiruo and Li, Xiaoxi and Zhu, Yutao and Dou, Zhicheng and Nie, Jian-Yun},
  journal={ACM Transactions on Information Systems},
  volume={43},
  number={6},
  pages={1--50},
  year={2025},
  publisher={ACM New York, NY}
}

@inproceedings{mo2023convgqr,
  title={ConvGQR: Generative Query Reformulation for Conversational Search},
  author={Mo, Fengran and Mao, Kelong and Zhu, Yutao and Wu, Yihong and Huang, Kaiyu and Nie, Jian-Yun},
  booktitle={Proceedings of the 61st Annual Meeting of the Association for Computational Linguistics (Volume 1: Long Papers)},
  pages={4998--5012},
  year={2023}
}

@article{yu2025dapo,
  title={Dapo: An open-source llm reinforcement learning system at scale},
  author={Yu, Qiying and Zhang, Zheng and Zhu, Ruofei and Yuan, Yufeng and Zuo, Xiaochen and Yue, Yu and Dai, Weinan and Fan, Tiantian and Liu, Gaohong and Liu, Lingjun and others},
  journal={arXiv preprint arXiv:2503.14476},
  year={2025}
}

@inproceedings{huang-etal-2025-crmarena,
    title = "{CRMA}rena: Understanding the Capacity of {LLM} Agents to Perform Professional {CRM} Tasks in Realistic Environments",
    author = "Huang, Kung-Hsiang  and
      Prabhakar, Akshara  and
      Dhawan, Sidharth  and
      Mao, Yixin  and
      Wang, Huan  and
      Savarese, Silvio  and
      Xiong, Caiming  and
      Laban, Philippe  and
      Wu, Chien-Sheng",
    editor = "Chiruzzo, Luis  and
      Ritter, Alan  and
      Wang, Lu",
    booktitle = "Proceedings of the 2025 Conference of the Nations of the Americas Chapter of the Association for Computational Linguistics: Human Language Technologies (Volume 1: Long Papers)",
    month = apr,
    year = "2025",
    address = "Albuquerque, New Mexico",
    publisher = "Association for Computational Linguistics",
    url = "https://aclanthology.org/2025.naacl-long.194/",
    doi = "10.18653/v1/2025.naacl-long.194",
    pages = "3830--3850",
    ISBN = "979-8-89176-189-6"
}

@article{huang2025crmarena-pro,
  title={Crmarena-pro: Holistic assessment of llm agents across diverse business scenarios and interactions},
  author={Huang, Kung-Hsiang and Prabhakar, Akshara and Thorat, Onkar and Agarwal, Divyansh and Choubey, Prafulla Kumar and Mao, Yixin and Savarese, Silvio and Xiong, Caiming and Wu, Chien-Sheng},
  journal={arXiv preprint arXiv:2505.18878},
  year={2025}
}

@article{shi2025taskcraft,
  title={Taskcraft: Automated generation of agentic tasks},
  author={Shi, Dingfeng and Cao, Jingyi and Chen, Qianben and Sun, Weichen and Li, Weizhen and Lu, Hongxuan and Dong, Fangchen and Qin, Tianrui and Zhu, King and Liu, Minghao and others},
  journal={arXiv preprint arXiv:2506.10055},
  year={2025}
}

@misc{yuan2023scalingrelationshiplearningmathematical,
      title={Scaling Relationship on Learning Mathematical Reasoning with Large Language Models}, 
      author={Zheng Yuan and Hongyi Yuan and Chengpeng Li and Guanting Dong and Keming Lu and Chuanqi Tan and Chang Zhou and Jingren Zhou},
      year={2023},
      eprint={2308.01825},
      archivePrefix={arXiv},
      primaryClass={cs.CL},
      url={https://arxiv.org/abs/2308.01825}, 
}

@misc{grattafiori2024llama3herdmodels,
      title={The Llama 3 Herd of Models}, 
      author={Aaron Grattafiori et.al},
      year={2024},
      eprint={2407.21783},
      archivePrefix={arXiv},
      primaryClass={cs.AI},
      url={https://arxiv.org/abs/2407.21783}, 
}

@inproceedings{zheng2024llamafactory,
  title={LlamaFactory: Unified Efficient Fine-Tuning of 100+ Language Models},
  author={Yaowei Zheng and Richong Zhang and Junhao Zhang and Yanhan Ye and Zheyan Luo and Zhangchi Feng and Yongqiang Ma},
  booktitle={Proceedings of the 62nd Annual Meeting of the Association for Computational Linguistics (Volume 3: System Demonstrations)},
  address={Bangkok, Thailand},
  publisher={Association for Computational Linguistics},
  year={2024},
  url={http://arxiv.org/abs/2403.13372}
}

@article{fang2025memp,
  title={Memp: Exploring Agent Procedural Memory},
  author={Fang, Runnan and Liang, Yuan and Wang, Xiaobin and Wu, Jialong and Qiao, Shuofei and Xie, Pengjun and Huang, Fei and Chen, Huajun and Zhang, Ningyu},
  journal={arXiv preprint arXiv:2508.06433},
  year={2025}
}

@article{tang2025agent,
  title={Agent kb: Leveraging cross-domain experience for agentic problem solving},
  author={Tang, Xiangru and Qin, Tianrui and Peng, Tianhao and Zhou, Ziyang and Shao, Daniel and Du, Tingting and Wei, Xinming and Xia, Peng and Wu, Fang and Zhu, He and others},
  journal={arXiv preprint arXiv:2507.06229},
  year={2025}
}

@article{yao2022webshop,
  title={Webshop: Towards scalable real-world web interaction with grounded language agents},
  author={Yao, Shunyu and Chen, Howard and Yang, John and Narasimhan, Karthik},
  journal={Advances in Neural Information Processing Systems},
  volume={35},
  pages={20744--20757},
  year={2022}
}

@inproceedings{drouin2024workarena,
  title={WorkArena: how capable are web agents at solving common knowledge work tasks?},
  author={Drouin, Alexandre and Gasse, Maxime and Caccia, Massimo and Laradji, Issam H and Verme, Manuel Del and Marty, Tom and Vazquez, David and Chapados, Nicolas and Lacoste, Alexandre},
  booktitle={Proceedings of the 41st International Conference on Machine Learning},
  pages={11642--11662},
  year={2024}
}

@inproceedings{
workbench,
title={WorkBench: a Benchmark Dataset for Agents in a Realistic Workplace Setting},
author={Olly Styles and Sam Miller and Patricio Cerda-Mardini and Tanaya Guha and Victor Sanchez and Bertie Vidgen},
booktitle={First Conference on Language Modeling},
year={2024},
url={https://openreview.net/forum?id=4HNAwZFDcH}
}

@misc{yao2024tau,
      title={$\tau$-bench: A Benchmark for Tool-Agent-User Interaction in Real-World Domains}, 
      author={Shunyu Yao and Noah Shinn and Pedram Razavi and Karthik Narasimhan},
      year={2024},
      eprint={2406.12045},
      archivePrefix={arXiv},
      primaryClass={cs.AI},
      url={https://arxiv.org/abs/2406.12045}, 
}

@article{long-term-memory-survey,
  title={Human-inspired Perspectives: A Survey on AI Long-term Memory},
  author={He, Zihong and Lin, Weizhe and Zheng, Hao and Zhang, Fan and Jones, Matt W and Aitchison, Laurence and Xu, Xuhai and Liu, Miao and Kristensson, Per Ola and Shen, Junxiao},
  journal={arXiv preprint arXiv:2411.00489},
  year={2024}
}

@inproceedings{zhong2024memorybank,
  title={Memorybank: Enhancing large language models with long-term memory},
  author={Zhong, Wanjun and Guo, Lianghong and Gao, Qiqi and Ye, He and Wang, Yanlin},
  booktitle={Proceedings of the AAAI Conference on Artificial Intelligence},
  volume={38},
  number={17},
  pages={19724--19731},
  year={2024}
}

@inproceedings{memoryllm,
  author={Yu Wang and Yifan Gao and Xiusi Chen and Haoming Jiang and Shiyang Li and Jingfeng Yang and Qingyu Yin and Zheng Li and Xian Li and Bing Yin and Jingbo Shang and Julian J. McAuley},
  title={MEMORYLLM: Towards Self-Updatable Large Language Models},
  year={2024},
  cdate={1704067200000},
  url={https://openreview.net/forum?id=p0lKWzdikQ},
  booktitle={ICML},
}

@inproceedings{
agent-workflow-memory,
title={Agent Workflow Memory},
author={Zora Zhiruo Wang and Jiayuan Mao and Daniel Fried and Graham Neubig},
booktitle={Forty-second International Conference on Machine Learning},
year={2025},
url={https://openreview.net/forum?id=NTAhi2JEEE}
}

@article{chhikara2025mem0,
  title={Mem0: Building production-ready ai agents with scalable long-term memory},
  author={Chhikara, Prateek and Khant, Dev and Aryan, Saket and Singh, Taranjeet and Yadav, Deshraj},
  journal={arXiv preprint arXiv:2504.19413},
  year={2025}
}

@article{guo2025deepseek,
  title={Deepseek-r1: Incentivizing reasoning capability in llms via reinforcement learning},
  author={Guo, Daya and Yang, Dejian and Zhang, Haowei and Song, Junxiao and Zhang, Ruoyu and Xu, Runxin and Zhu, Qihao and Ma, Shirong and Wang, Peiyi and Bi, Xiao and others},
  journal={arXiv preprint arXiv:2501.12948},
  year={2025}
}

@article{jin2025search,
  title={Search-r1: Training llms to reason and leverage search engines with reinforcement learning},
  author={Jin, Bowen and Zeng, Hansi and Yue, Zhenrui and Yoon, Jinsung and Arik, Sercan and Wang, Dong and Zamani, Hamed and Han, Jiawei},
  journal={arXiv preprint arXiv:2503.09516},
  year={2025}
}

@article{feng2025retool,
  title={Retool: Reinforcement learning for strategic tool use in llms},
  author={Feng, Jiazhan and Huang, Shijue and Qu, Xingwei and Zhang, Ge and Qin, Yujia and Zhong, Baoquan and Jiang, Chengquan and Chi, Jinxin and Zhong, Wanjun},
  journal={arXiv preprint arXiv:2504.11536},
  year={2025}
}

@article{tao2025webshaper,
  title={Webshaper: Agentically data synthesizing via information-seeking formalization},
  author={Tao, Zhengwei and Wu, Jialong and Yin, Wenbiao and Zhang, Junkai and Li, Baixuan and Shen, Haiyang and Li, Kuan and Zhang, Liwen and Wang, Xinyu and Jiang, Yong and others},
  journal={arXiv preprint arXiv:2507.15061},
  year={2025}
}

@inproceedings{
wang2025openhands,
title={OpenHands: An Open Platform for {AI} Software Developers as Generalist Agents},
author={Xingyao Wang and Boxuan Li and Yufan Song and Frank F. Xu and Xiangru Tang and Mingchen Zhuge and Jiayi Pan and Yueqi Song and Bowen Li and Jaskirat Singh and Hoang H. Tran and Fuqiang Li and Ren Ma and Mingzhang Zheng and Bill Qian and Yanjun Shao and Niklas Muennighoff and Yizhe Zhang and Binyuan Hui and Junyang Lin and Robert Brennan and Hao Peng and Heng Ji and Graham Neubig},
booktitle={The Thirteenth International Conference on Learning Representations},
year={2025},
url={https://openreview.net/forum?id=OJd3ayDDoF}
}

@article{deng2023mind2web,
  title={Mind2web: Towards a generalist agent for the web},
  author={Deng, Xiang and Gu, Yu and Zheng, Boyuan and Chen, Shijie and Stevens, Sam and Wang, Boshi and Sun, Huan and Su, Yu},
  journal={Advances in Neural Information Processing Systems},
  volume={36},
  pages={28091--28114},
  year={2023}
}

@article{zhao2025sciarena,
  title={SciArena: An Open Evaluation Platform for Foundation Models in Scientific Literature Tasks},
  author={Zhao, Yilun and Zhang, Kaiyan and Hu, Tiansheng and Wu, Sihong and Bras, Ronan Le and Anderson, Taira and Bragg, Jonathan and Chang, Joseph Chee and Dodge, Jesse and Latzke, Matt and others},
  journal={arXiv preprint arXiv:2507.01001},
  year={2025}
}

@misc{areal-rl-framework,
      title={AReaL: A Large-Scale Asynchronous Reinforcement Learning System for Language Reasoning}, 
      author={Wei Fu and Jiaxuan Gao and Xujie Shen and Chen Zhu and Zhiyu Mei and Chuyi He and Shusheng Xu and Guo Wei and Jun Mei and Jiashu Wang and Tongkai Yang and Binhang Yuan and Yi Wu},
      year={2025},
      eprint={2505.24298},
      archivePrefix={arXiv},
      primaryClass={cs.LG},
      url={https://arxiv.org/abs/2505.24298}, 
}

@misc{resum,
      title={ReSum: Unlocking Long-Horizon Search Intelligence via Context Summarization}, 
      author={Xixi Wu and Kuan Li and Yida Zhao and Liwen Zhang and Litu Ou and Huifeng Yin and Zhongwang Zhang and Yong Jiang and Pengjun Xie and Fei Huang and Minhao Cheng and Shuai Wang and Hong Cheng and Jingren Zhou},
      year={2025},
      eprint={2509.13313},
      archivePrefix={arXiv},
      primaryClass={cs.CL},
      url={https://arxiv.org/abs/2509.13313}, 
}

@inproceedings{adacqr,
    title = "{A}da{CQR}: Enhancing Query Reformulation for Conversational Search via Sparse and Dense Retrieval Alignment",
    author = "Lai, Yilong  and
      Wu, Jialong  and
      Zhang, Congzhi  and
      Sun, Haowen  and
      Zhou, Deyu",
    editor = "Rambow, Owen  and
      Wanner, Leo  and
      Apidianaki, Marianna  and
      Al-Khalifa, Hend  and
      Eugenio, Barbara Di  and
      Schockaert, Steven",
    booktitle = "Proceedings of the 31st International Conference on Computational Linguistics",
    month = jan,
    year = "2025",
    address = "Abu Dhabi, UAE",
    publisher = "Association for Computational Linguistics",
    url = "https://aclanthology.org/2025.coling-main.515/",
    pages = "7698--7720"
}

@inproceedings{adarewriter,
    title = "{A}da{R}ewriter: Unleashing the Power of Prompting-based Conversational Query Reformulation via Test-Time Adaptation",
    author = "Lai, Yilong  and
      Wu, Jialong  and
      Wang, Zhenglin  and
      Zhou, Deyu",
    editor = "Christodoulopoulos, Christos  and
      Chakraborty, Tanmoy  and
      Rose, Carolyn  and
      Peng, Violet",
    booktitle = "Proceedings of the 2025 Conference on Empirical Methods in Natural Language Processing",
    month = nov,
    year = "2025",
    address = "Suzhou, China",
    publisher = "Association for Computational Linguistics",
    url = "https://aclanthology.org/2025.emnlp-main.193/",
    doi = "10.18653/v1/2025.emnlp-main.193",
    pages = "3889--3905",
    ISBN = "979-8-89176-332-6"
}
\clearpage
\appendix

\section*{Appendix}

\section{Long-term Memory Details}\label{appendix:long-term-memory}
\begin{algorithm}[htbp]
\caption{Long-term Memory Module with Offline Memory Consolidation}
\label{algorithm:memory}
\KwIn{Query $Q$, Indexed Memory Pool $\mathcal{M}$, Threshold $\tau$, System Prompt $P$}
\KwOut{Execution Trajectory $\mathcal{S}$}

\algobox{\textbf{Online Phase: Memory-Guided Inference}}

$\triangleright$ Retrieve Top-$1$ memory candidate $m$ and similarity score $\phi$ from $\mathcal{M}$ using $Q$\;

\eIf{$\phi \ge \tau$}{
    $P \leftarrow \textsc{ContextAugment}(P, m)$\; \color{gray}{\tcp{{Incorporating memory as contextual information.}}}
}{
    \color{gray}{\tcp{{Proceed without external memory}}}
}

$\mathcal{S} \leftarrow \textsc{Solve}(Q \mid P; \theta)$\;

\algobox{\textbf{Offline Phase: Memory Consolidation}}

\If{$\phi < \tau$}{
    $\tilde{\mathcal{S}} \leftarrow \textsc{Solve}(\mathcal{Q;H}_{adv})$\; \textcolor{gray}{\tcp{{Try to solve with consolidation model $\mathcal{H}_{adv}$}}}
    $\mathcal{E} \leftarrow \textsc{BuildMemory}(\tilde{\mathcal{S}})$\; \textcolor{gray}{\tcp{Obtain memory unit $\mathcal{E}$ based on the trajectory $\tilde{\mathcal{S}}$}}
    $r \leftarrow \textsc{Check}(\mathcal{E})$\;
    \If{$r = \texttt{accept}$}{
        $\mathcal{M} \leftarrow \mathcal{M} \cup \{\mathcal{E}\}$\;
    }
}

\Return $\mathcal{S}$\;
\end{algorithm}

\subsection{Memory Indexing}\label{appendix:memory_generation}
To generate the long-term memory unit $\mathcal{E}$, we use \texttt{o3-mini} as the advanced reasoning model.
After obtaining the trajectories $\mathcal{H}_{\text{adv}}$, we use \texttt{GPT-4.1} with a prompt described in Table~\ref{prompt_for_generating_guideline} to generate the guideline for the given task. 

    
    
    


    
    

\subsection{Evaluation Protocol for Memory Module}~\label{appendix:evaluate_protocol}
During evaluation on CRMArena-Pro, the long-term memory module is allowed to be updated via the offline consolidation procedure described in Section~\ref{sec:memory}, reflecting a continual deployment setting.

To ensure reproducibility and reduce potential order effects, test queries are randomly permuted and evaluated over three independent runs, and performance is reported as the average across these runs.
We note that this protocol differs from a strictly frozen-memory evaluation, as memory updates are permitted within each run. 
Nevertheless, such updates occur infrequently (approximately 3\% on average in the B2B setting), and the results remain stable across different random orderings.

\section{Implementation Details}\label{appendix:implementation_details}
In our RL training, we set the rollout size $G=16$, with $\varepsilon_{\text {low }}$ at 0.2 and $\varepsilon_{\text {high}}$ at 0.28. 
Moreover, to support stable model training during RL, we employ the official, locally deployable SQLite database\footnote{\url{https://github.com/SalesforceAIResearch/CRMArena/tree/main/local_data}} for the SQL tool. 
In the case of the SOSL tool, we follow the original configuration, which relies on invoking the Salesforce API.
The threshold $\tau$ in Algorithm~\ref{algorithm:memory} is set to $0.7$.
During inference, we set the temperature to 0.2 and limited the maximum number of interaction turns to 20.
For the subsequent main experiments, the results of the OpenAI and Gemini series models are taken directly from the original paper~\cite{huang2025crmarena-pro}. 
We have also included results from the Qwen3-235B-A22B-Instruct and Kimi-K2 model, using the ReAct paradigm, for comparison.
All experiments are conducted on 2 nodes equipped with 8 x Nvidia H20.

\section{Details of CRMArena-Pro}\label{appendix:crmarena-details}
\begin{table}[t]
\centering
\small
\begin{tabularx}{\columnwidth}{lcc}
\toprule
\multirow{2}{*}{\textbf{Tasks}} & \multicolumn{2}{c}{\textbf{Amount}} \\
 & \textbf{B2B} & \textbf{B2C} \\
\midrule

\emph{Workflow Execution} \\
\quad Service Case Routing & 100 & 100 \\
\quad Sales Lead Routing & 100 & 100 \\

\midrule
\emph{Policy Compliance} \\
\quad Invalid Configuration Identification & 100 & 100 \\
\quad Solution Violation Identification & 100 & 100 \\
\quad Lead Qualification & 100 & 100 \\
\quad Quote Approval & 100 & 100 \\

\midrule
\emph{Information Retrieval \& Reasoning} \\
\quad Knowledge Question Answering & 100 & 100 \\
\quad Sales Insight Mining & 100 & 100 \\
\quad Wrong Stage Rectification & 100 & 100 \\
\quad Activity Priority Understanding & 100 & 100 \\
\quad Named Entity Disambiguation & 100 & 100 \\

\midrule
\emph{Structured Data \& Computation} \\
\quad Handle Time Understanding & 100 & 100 \\
\quad Transfer Count Understanding & 100 & 100 \\
\quad Monthly Trend Analysis & 100 & 100 \\
\quad Conversion Rate Comprehension & 100 & 100 \\
\quad Best Region Identification & 100 & 100 \\
\quad Sales Volume Understanding & 100 & 100 \\
\quad Sales Cycle Understanding & 100 & 100 \\
\quad Conversion Rate Comprehension & 100 & 100 \\

\midrule
\textbf{Total(19 Tasks)} & \textbf{1900} & \textbf{1900} \\
\bottomrule
\end{tabularx}
\caption{Detailed dataset statistics of CRMArena-Pro.}
\label{tab:task_completion}
\end{table}

CRMArena-Pro comprises four categories of tasks and two types of scenarios(B2B and B2C), enabling a comprehensive evaluation of agent performance on business-related tasks.

We provide detailed dataset statistics of CRMArena-Pro in Table~\ref{tab:task_completion}, and here is the detailed information for the four categories:

\begin{itemize}
    \item \textbf{Database Querying \& Numerical Computation(Data)}: This skill assesses an agent's ability to formulate structured queries and perform numerical calculations on the retrieved data.
    \item \textbf{Information Retrieval \& Textual Reasoning(Text)}: This involves processing and reasoning over unstructured text from sources such as knowledge bases or call transcripts to extract insights.
    \item \textbf{Workflow Execution(Workflow)}: This evaluates the agent's capacity to follow pre-defined business processes and execute actions based on specific rules.
    \item \textbf{Policy Compliance(Policy)}: This tests an agent's ability to verify whether business operations adhere to established company policies and regulations.
\end{itemize}

\section{Synthesis Data Construction}\label{appendix:synthesis_data}

\subsection{Prompt for Complex Queries Generation}
We provide the prompt used to generate the complex queries in Table~\ref{prompt_init_complex_query} and Table~\ref{prompt_update_complex_query}.

\subsection{Complex Synthesis Data Examples}\label{appendix:complex_synthesis_data_example}
Here are some examples of complex synthesis data:
\begin{enumerate}
    \item Which live chat transcript documents a case where a Portland-based customer, previously known for resolving product and sizing complaints in 2022, reached out on November 30, 2023, to request an immediate exchange after receiving a blue hoodie instead of black running shoes, and successfully initiated the return process with photo documentation through interactions involving three different support agents? Return only the chat transcript's ID. If no such records exist, return 'None'.
    \item During the second week of January 2022, a customer with an account connected to a contact named Isabella Adams, who has an email address of isabella.morgan@example.com and a shipping address in Austin, Texas, placed an order that was activated and used a price book featuring the promotion 'Start 2022 with unbeatable deals on sports gear.' Which price book was referenced in this transaction? Return only the price book's id. If no such records exist, return 'None'.
    \item Identify the durable and waterproof item, ideal for harsh weather conditions, that was added as a new selection by the customer whose previous order item was related to outdoor hiking, after a warranty issue with defective equipment was resolved in Portland and resulted in a successful transaction activation on March 19, 2022. What is the unique catalog reference for that item within the product listings? Return only the product's ID. If no such records exist, return 'None'.
\end{enumerate}

\begin{table*}
\begin{tcolorbox}[colback=white!95!gray,colframe=gray!50!black,title=Prompt for Generating Guideline]
\begin{lstlisting}[breaklines=true, xleftmargin=0pt, breakindent=0pt, columns=fullflexible]
Please provide a concise guideline for the following execution log.
Break down the guideline into individual steps, and for each step, briefly describe what action was taken (note that it must be report the tool's name explicitly) and its outcome. 
Ensure the summary is clear, structured, and captures all key actions in the order they occurred.

## Key Requirements
1. Focus exclusively on behavioral improvements derived from similar task patterns and experience.
2. Format output strictly as:
    1. [Specific suggestion 1]
    2. [Specific suggestion 2]
    ...
3. No headings, explanations, or markdown.

## Detailed Schema Information
{schema_information}

## Execution logs:
{successful_trajectories}
\end{lstlisting}
\end{tcolorbox}
\caption{The prompt used to generating the guideline of memory unit $\mathcal{E}$.}
\label{prompt_for_generating_guideline}
\end{table*}

\begin{table*}[!htbp]
\centering
\begin{minipage}{0.9\linewidth}
\begin{tcolorbox}[colback=white!95!gray,colframe=gray!50!black]
\textbf{Key:} Can we qualify this lead based on their recent discussions?...

\textbf{Value:}
\begin{lstlisting}[breaklines=true, xleftmargin=0pt, breakindent=0pt, columns=fullflexible]
1. Always use the execute tool to pull every relevant transcript and pricing/knowledge record before judging any BANT factor, ensuring no critical data is missed.
2. After data retrieval, perform an explicit cost-vs-budget calculation inside the analysis step to avoid incorrect 'Budget' conclusions.
3. When Authority is not clearly confirmed in the transcript, trigger an additional execute search (e.g., manager titles, purchasing roles) before deciding it is fulfilled or unmet.  
4. Parse Timeline details quantitatively (days, dates) rather than qualitatively; if uncertain, run another execute query for scheduling constraints to improve accuracy.  
5. Maintain a consistent workflow order -> execute -> analyze -> respond to increase reliability and reproducibility across similar tasks.
\end{lstlisting}
\end{tcolorbox}
\caption{An example of Memory Unit $\mathcal{E}$, where the \textbf{Key} is the indexed query used to match similar queries, and \textbf{Value} is the corresponding guideline.
}
\vspace{-8mm}
\label{table:memory_example}
\end{minipage}
\end{table*}

\begin{table*}
\begin{tcolorbox}[colback=white!95!gray,colframe=gray!50!black,title=Prompt for Initialize Seed Query]
\begin{lstlisting}[breaklines=true, xleftmargin=0pt, breakindent=0pt, columns=fullflexible]
You are a synthesis data generator. I will give you two schema and two corresponding data records. The two records are connected by a foreign key relationship. Your task is to generate a complex natural language question that includes the information from the records implicitly. Apart from this, the question's answer should be the id I provided with you. The question should be complex and not directly related to the id. 

Input/Output Specifications:
Input:
- Source Schema: A description of the source schema.
- Source Record: A record from the source schema.
- Target Schema: A description of the target schema.
- Target Record: A record from the target schema.
- Connect Key: The connect relationship between the two records.
- Answer: Known correct answer.

Output:
- Must be in strict JSON format: {"Q": "generated question"}
- No explanations or extra fields allowed

Q must satisfy:
1. Be a complete natural language question
2. Allow deriving answer A by using the source and target records and connect key

Question Generation Principles:
1. Exact correspondence - Each question must fully base on the original conclusion, with the answer being its core content.
2. Derivability - The original conclusion must be directly derivable from the question and be the only correct answer.
3. Self-containment - Questions must be complete and independent, not relying on external references or unspecified context.
4. Information hiding - Do not reveal specific sources or data paths, but can include search hints.
5. Specificity and clarity - Questions should include details like specific times to ensure unique answers.
6. Single question - Generate only one question per conclusion.
7. If the conclusion can only be obtained from input content, include hints via data source identifiers in the question.
8. Language consistency - The language of each question must be the same as the conclusion's language.
\end{lstlisting}
\end{tcolorbox}
\caption{Prompt for Generating Seed Query}
\label{prompt_init_complex_query}
\end{table*}

\begin{table*}
\begin{tcolorbox}[colback=white!95!gray,colframe=gray!50!black,title=Prompt for Initialize Seed Query]
\begin{lstlisting}[breaklines=true, xleftmargin=0pt, breakindent=0pt, columns=fullflexible]
You are a synthesis data generator. I will give you an existing question and its question. Apart from this, I will also give you a schema and a record. The schema and record includes a foreign key exactly the same as the answer of the giving question. Your task is to generate a more complex question that merges the information from the record and the current given record implicitly. Apart from this, the question's answer should be the id I provided with you. The question should be complex and not directly related to the ID. And the question should not include explicit information such as id of the record or the name of the user.

Input/Output Specifications:
Input:
- Existing Question: A complex natural language question that has been previously generated.
- Existing Question's Answer: The answer to the existing question.
- Target Schema: A description of the target schema.
- Target Record: A record from the target schema.
- Connect Key: The connect relationship between the two records.
- Answer: Known correct answer.

Output:
- Must be in strict JSON format: {"Q": "generated question"}
- No explanations or extra fields allowed

Q must satisfy:
1. Be a complete natural language question
2. Allow deriving answer A by using the source and target records and the connect key

Question Generation Principles:
1. Exact correspondence - Each question must be fully based on the original conclusion, with the answer being its core content.
2. Derivability - The original conclusion must be directly derivable from the question and be the only correct answer.
3. Self-containment - Questions must be complete and independent, not relying on external references or unspecified context.
4. Information hiding - Do not reveal specific sources or data paths, but can include search hints.
5. Specificity and clarity - Questions should include details like specific times to ensure unique answers.
6. Single question - Generate only one question per conclusion.
7. If the conclusion can only be obtained from input content, include hints via data source identifiers in the question.
8. Language consistency - The language of each question must be the same as the conclusion's language.
\end{lstlisting}
\end{tcolorbox}
\caption{Prompt for Generating Deeper Challenge Queries}
\label{prompt_update_complex_query}
\end{table*}

\section{Case Study of CRMWeaver}\label{appendix:case_study}
\subsection{Long-term Memory Example}
Table~\ref{table:memory_example} provide a example of memory unit $\mathcal{E}$.

\subsection{Full Trajectories Example}
Table~\ref{trajectories_example} is a full trajectories example of our proposed agent.

\onecolumn
\begin{tcolorbox}[
breakable,
colback=white!95!gray,
colframe=gray!50!black,
title=Trajectories Example on CRMArena-Pro,
width=\textwidth,
break at=1.0\textheight,  
]
\tcbox[
    colback=yellow!50,
    colframe=yellow,
    arc=0pt,
    height=0.5cm,
    halign=center,
    valign=center,
    fontupper=\bfseries
]{Question:}
\begin{lstlisting}[breaklines=true, xleftmargin=0pt, breakindent=0pt, columns=fullflexible]
Which agent ought to be designated to handle this lead? Return only the ID of the agent.
## Lead Routing Policy
The lead routing policy determines the best agent to assign the given new lead based on the following criteria:
- Territory Match: First, find the territory that matches the lead's region by comparing it with the territory description (comma-separated list of states/regions).
- Quote Success: Among agents associated with the matching territory, select those with the highest number of quotes with status "Accepted" (find the agent ID through the associated opportunity ID).
- Workload Balance: If there's still a tie, select the agent with the fewest unconverted (open) leads
## Lead's region: WI
\end{lstlisting}

\tcbox[
    colback=yellow!50,
    colframe=yellow,
    arc=0pt,
    height=0.5cm,
    halign=center,
    valign=center,
    fontupper=\bfseries,
]{Step 1}

\textbf{\textcolor{orange}{<think>}}

\textcolor{orange}{I need to find the territory that matches the lead's region (WI) by checking the Territory2 object. I will look for a territory with "WI" in its Description field.}

\textbf{\textcolor{orange}{</think>}}\\

\textbf{\textcolor{cyan}{<tool\_call>}}

\texttt{\textcolor{cyan}{\{"name": "execute", "arguments": \{"query": "SELECT Id, Name FROM Territory2 WHERE Description LIKE '\%WI\%"\}\}}}

\textbf{\textcolor{cyan}{</tool\_call>}}\\

\textbf{<tool\_response>}

\texttt{Observation: [('0MIWt0000007xzROAQ', 'US-Great Lakes')]}

\textbf{</tool\_response>}

\tcbox[
    colback=yellow!50,
    colframe=yellow,
    arc=0pt,
    height=0.5cm,
    halign=center,
    valign=center,
    fontupper=\bfseries
]{Step 2}

\textbf{\textcolor{orange}{<think>}}

\textcolor{orange}{I have found the territory that matches the lead's region (WI), which is "US-Great Lakes" with ID \texttt{"0MIWt0000007xzROAQ"}. Now, I need to find the agents associated with this territory by checking the UserTerritory2Association object.}

\textbf{\textcolor{orange}{</think>}}\\

\textbf{\textcolor{cyan}{<tool\_call>}}

\texttt{\textcolor{cyan}{\{"name": "execute", "arguments": \{"query": "SELECT UserId FROM UserTerritory2Association WHERE Territory2Id = '0MIWt0000007xzROAQ'"\}\}}}

\textbf{\textcolor{cyan}{</tool\_call>}}\\

\textbf{<tool\_response>}

\texttt{Observation: [('005Wt000003NHpdIAG',), ...]}

\textbf{</tool\_response>}

\tcbox[
    colback=yellow!50,
    colframe=yellow,
    arc=0pt,
    height=0.5cm,
    halign=center,
    valign=center,
    fontupper=\bfseries
]{Step 3}

\textbf{\textcolor{orange}{<think>}}

\textcolor{orange}{I have found the list of agents associated with the territory "US-Great Lakes" (WI). Now, I need to determine which of these agents has the highest number of quotes with status "Accepted" by checking the Quote and QuoteLineItem objects.}

\textbf{\textcolor{orange}{</think>}}\\

\textbf{\textcolor{cyan}{<tool\_call>}}

\textcolor{cyan}{\texttt{\{"name": "execute", "arguments": \{"query": "SELECT OwnerId, COUNT(*) AS accepted\_quotes FROM Quote WHERE Status = 'Accepted' AND OwnerId IN ('005Wt000003NHpdIAG', $\cdots$, '005Wt000003NJmcIAG') GROUP BY OwnerId ORDER BY accepted\_quotes DESC"\}\}}}

\textbf{\textcolor{cyan}{</tool\_call>}}\\

\textbf{<tool\_response>}

Observation: Error: no such column: OwnerId

\textbf{</tool\_response>}

\tcbox[
    colback=yellow!50,
    colframe=yellow,
    arc=0pt,
    height=0.5cm,
    halign=center,
    valign=center,
    fontupper=\bfseries
]{Step 4}

\textbf{\textcolor{orange}{<think>}}

\textcolor{orange}{The error occurred because the OwnerId field does not exist directly on the Quote object. Instead, the agent (OwnerId) is associated with the Opportunity object, and the Quote is linked to that Opportunity via OpportunityId. I need to check the Opportunity object to find the OwnerId (agent) associated with each Quote.}

\textbf{\textcolor{orange}{</think>}}\\

\textbf{\textcolor{cyan}{<tool\_call>}}

\textcolor{cyan}{\texttt{\{"name": "execute", "arguments": \{"query": "SELECT o.OwnerId, COUNT(*) AS accepted\_quotes FROM Quote q JOIN Opportunity o ON q.OpportunityId = o.Id WHERE q.Status = 'Accepted' AND o.OwnerId IN ('005Wt000003NHpdIAG', $\cdots$, '005Wt000003NJmcIAG') GROUP BY o.OwnerId ORDER BY accepted\_quotes DESC"\}\}}}

\textbf{\textcolor{cyan}{</tool\_call>}}\\

\textbf{<tool\_response>}

\texttt{Observation: [('\texttt{005Wt000003NIowIAG}', 5), ('005Wt000003NJmcIAG', 4),$\cdots$, ('005Wt000003NDXaIAO', 1)]}

\textbf{</tool\_response>}

\tcbox[
    colback=yellow!50,
    colframe=yellow,
    arc=0pt,
    width=\linewidth, 
    height=0.5cm,
    halign=center,
    valign=center,
    fontupper=\bfseries
]{Step 5}

\textbf{\textcolor{orange}{<think>}}

\textcolor{orange}{The agent with ID \texttt{'005Wt000003NIowIAG'} has the highest number of accepted quotes (5), which is more than any other agent in the list. Since this agent has the highest quote success, they should be allocated to the lead according to the routing policy.}

\textbf{\textcolor{orange}{</think>}}\\

\textbf{\textcolor{red}{<answer>}}

\textcolor{red}{\texttt{005Wt000003NIowIAG}}

\textbf{\textcolor{red}{</answer>}}
\end{tcolorbox}
\captionof{table}{A Trajectories Example on CRMArena-Pro.}
\label{trajectories_example}

\end{document}